\documentclass[runningheads]{llncs}

\usepackage[T1]{fontenc}

\usepackage{newtxtext}
\usepackage[varvw]{newtxmath}

\usepackage{microtype}
\usepackage{graphicx}
\usepackage{subcaption}
\usepackage{booktabs}
\usepackage{multirow}
\usepackage{hyperref}
\usepackage{colortbl}
\usepackage{arydshln}
\usepackage{orcidlink}
\usepackage{marvosym} 

\usepackage{amsmath,amssymb,amsfonts}
\usepackage{mathtools}

\usepackage{algorithm}
\usepackage{algorithmic}

\usepackage[capitalize,noabbrev]{cleveref}
\usepackage{xspace}

\newcommand{\doa}{\textsc{DOA}\xspace}
\newcommand{\igdoa}{\textsc{IG-DOA}\xspace}
\newcommand{\igska}{\textsc{IG-SKA}\xspace}

\begin{document}

\title{Let the Neurons Die: \\Exploiting ReLU-Induced Model Degradation}

\titlerunning{Let the Neurons Die}

\author{
Kexin Li\inst{1}\textsuperscript{\ensuremath{\dagger}}\orcidlink{0000-0002-7911-3731}
\and
Wenjun Qiu\inst{1}\textsuperscript{\ensuremath{\dagger}}\Letter\orcidlink{0009-0006-2375-1722}
\and
Joshua Abraham\textsuperscript{\ensuremath{\ddagger}}\orcidlink{0009-0009-8576-7123}
\and
Aditi Maheshwari\textsuperscript{\ensuremath{\ddagger}}\orcidlink{0009-0009-2959-3552}
\and
David Lie\inst{1}\orcidlink{0000-0002-2000-6827}
}

\authorrunning{K. Li et al.}

\institute{
University of Toronto, Toronto, Ontario, Canada\\
\email{wenjun.qiu@mail.utoronto.ca}
}

\makeatletter
\renewcommand{\lastandname}{,}
\makeatother

\maketitle

\begingroup
\renewcommand{\thefootnote}{\ensuremath{\dagger}}
\footnotetext{Equal contribution.}
\renewcommand{\thefootnote}{\ensuremath{\ddagger}}
\footnotetext{Work done while at the University of Toronto.}
\endgroup

\begin{abstract}
Rectified linear unit (ReLU) networks can suffer from \textit{dying neurons}, where units with persistently negative pre-activations produce zero outputs, blocking gradients through their activations. To exploit this failure mode, we present three training-time availability attacks based on data ordering and poisoning. We begin with the basic dynamic data-ordering attack (\doa{}), which greedily constructs a training prefix by selecting the next example that minimizes the target layer's post-update weight sum, aiming to push ReLU units toward negative pre-activations without modifying training samples or labels. We then develop two poisoning attacks, \igdoa{} and \igska{}, which use gradient inversion to synthesize class-conditioned samples by matching reference gradients in adverse model states constructed through \textit{data ordering} or \textit{soft knockout}, respectively. Soft knockout rearranges weights across adjacent layers to concentrate negative contributions. On a fully connected ReLU network trained on MNIST, ordering $100$ of $60{,}000$ training examples reduces test accuracy from $96\%$ to $95\%$ after only five epochs. Adding $200$ poisoned samples from a single class reduces test accuracy to approximately $86$-$88\%$ after five epochs in most evaluated conditions, compared with approximately $96\%$ under clean training. These results demonstrate that ReLU-targeted data ordering and poisoning can impair learning without directly modifying the victim model's parameters. Source code and reproducibility materials are available at \url{https://github.com/Kexin6/let_the_neurons_die}.
\end{abstract}

\section{Introduction}
\label{sec:introduction}

Rectified linear units (ReLU), $\sigma(z)=\max(0,z)$, are commonly used in neural networks because of their computational simplicity and empirical effectiveness~\cite{glorot2011deep}. By its nature, ReLU has a zero derivative for negative inputs, which leads to a failure mode known as \textit{dying ReLU}. This occurs when a unit's activation remains negative across training samples and produces no gradient flow through its activation~\cite{lu2020dying}. In the case where many units become inactive, the network can lose representational capacity and become difficult to optimize. This behavior makes ReLU inactivity a potential target for \emph{training-time availability attacks}: attacks intended to reduce learning ability rather than induce a specific misclassification.

Prior work shows that an attacker can trigger this failure mode by rearranging initial weights while preserving their marginal statistics~\cite{grosse2020security}. However, this approach requires control over the model's initialization. Separately, data-ordering attacks show that an adversary can disrupt training by manipulating the sequence of otherwise unchanged examples, using loss-based scores to select their order~\cite{shumailov2021manipulating}. Qiu's survey also identifies optimization failures, including vanishing gradients and poor initialization, as potential directions for poisoning attacks~\cite{qiu2022survey}. Motivated by these observations, we investigate whether training-sample order and synthetic data can be used to target ReLU inactivity without directly modifying the victim model's parameters. We consider a white-box attacker with access to the model, loss function, and training data, and distinguish two forms of control. An order-only attacker can reorder existing examples without changing their contents or labels. A poisoning attacker can instead add synthetic, labelled examples to a training set that the learner shuffles. In the poisoning setting, the victim is independently initialized. Constructing adverse weights is therefore not enough: the attack must act through the added examples rather than rely on installing those weights or preserving an attacker-selected order.

To exploit ReLU's failure mode under these constraints, we separate \emph{adverse-state construction} from \emph{poison synthesis}. We first introduce a dynamic data-ordering attack (\doa{}), which greedily constructs an ordered prefix of existing training examples. At each position, it evaluates unused candidates by replaying the current prefix followed by each candidate from a common checkpoint, then selects the candidate that minimizes the target layer's post-update weight sum. This objective aims to encourage negative pre-activations in the targeted ReLU layer. The resulting prefix is placed before the remaining training examples, leaving the examples and labels unchanged. \doa{} serves both as a standalone ordering attack and as one route to constructing an adverse model state. We then develop two poisoning attacks, \igdoa{} and \igska{}, that repurpose gradient inversion for poison synthesis. \igdoa{} uses a model state reached through \doa{}, whereas \igska{} adapts the soft-knockout construction of Grosse et al.~\cite{grosse2020security}. Soft knockout rearranges weights across adjacent layers to concentrate negative contributions; in our poisoning pipeline, this construction is performed on a local model, not on the victim. At either constructed state, we use gradient inversion~\cite{geiping2020inverting} to optimize synthetic inputs whose gradients match class-specific reference updates under a cosine-based objective and an image prior. The resulting labelled samples are added to the independently initialized victim's shuffled training set.

We evaluate all three attacks on a fully connected ReLU network trained on MNIST over five epochs. Ordering $100$ of $60{,}000$ training examples reduces test accuracy from $95.93\%$ to $95.04\%$ at the end of this training budget. For the poisoning attacks, adding $200$ synthetic examples labelled as a single class leaves most evaluated conditions at approximately $86$-$88\%$ accuracy, compared with approximately $96\%$ under clean training. Some poisoning conditions recover within the evaluation period. These results demonstrate modest degradation from ordering alone and larger, class-dependent losses from gradient-inverted poisons, even when the augmented training set is shuffled.

In summary, this paper makes the following contributions:
\begin{enumerate}
    \item \textbf{ReLU-targeted data ordering.} We introduce \doa{}, a greedy ordering attack that minimizes a targeted layer's post-update weight sum while leaving training examples and labels unchanged.
    \item \textbf{Availability poisoning through gradient inversion.} We develop \igdoa{} and \igska{}, which synthesize poisoned training samples by matching reference gradients at model states constructed through data ordering or adapted soft knockout.
    \item \textbf{A proof-of-concept evaluation on MNIST.} We quantify accuracy degradation across ordered-prefix lengths and class-conditioned poisoning settings, showing that the generated poisons can impair learning under shuffled training and independent victim initialization.
\end{enumerate}

\section{Background and Related Work}
\label{sec:related}

\paragraph{ReLU failure modes.}
Rectified linear units (ReLU) make deep neural networks easier and faster to train, partially because they support sparse activations and simple optimization~\cite{glorot2011deep}. The same design, however, also creates a well-known weakness: for negative pre-activations, the derivative becomes zero. As a result, some units become deactivated and may never recover during training, a phenomenon referred to as ``dying ReLU''~\cite{lu2020dying}. Grosse et al.\ show that this is not only an optimization issue but can also have security implications. By carefully permuting weights at initialization, an attacker can reshape how positive and negative contributions propagate across consecutive layers, causing downstream units to become inactive with high probability~\cite{grosse2020security}. Their attack assumes control over the initial weights. Our setting is different: we start from an ordinary initialization and evaluate whether the training sequence itself can drive the network toward a similarly harmful state. We then study whether gradients produced in that state can be used to construct malicious training samples.

\paragraph{Training-time attacks.}
Training-time poisoning attacks often introduce adversarially chosen examples and/or labels. Early work studied availability attacks mainly in convex learning settings, where poisoned data are optimized to degrade overall model performance~\cite{biggio2012poisoning,jagielski2018manipulating}. Other work considers clean-label attacks, which preserve apparently correct labels while steering the model towards specific errors~\cite{shafahi2018poison}. Defenses such as certified filtering can provide guarantees with bounded contamination, provided that the learning setting satisfies certain assumptions~\cite{steinhardt2017certified}. More recent attacks use gradient-based optimization, including gradient alignment, to make poisoning practical for deep networks~\cite{fowl2021adversarial}. Our goal is different from backdoor insertion or forcing a particular misclassification. We focus on availability: reducing the model's ability to learn effectively. 

\paragraph{Data ordering.}
The trajectory taken by Stochastic Gradient Descent (SGD) depends on the order in which training examples are presented, especially in non-convex models. Shumailov et al.\ showed that this dependence can be exploited through malicious batch ordering, reshuffling, and replacement. In some cases, even one adversarially arranged epoch can undo previous training progress or affect model behaviours~\cite{shumailov2021manipulating}. Their methods rank examples using loss-based scores computed on a target or surrogate model. Our approach uses a simpler criterion. The greedy ordering policy scores examples according to their effect on the sum of weights in a chosen layer, with the aim of pushing a ReLU layer towards more negative pre-activations.

\paragraph{Gradient inversion.}
Gradient inversion attacks were originally developed to recover private training inputs from gradients shared during training~\cite{zhu2019deep,geiping2020inverting,yin2021see}. For instance, Geiping et al. \ combine a cosine-based gradient matching objective with image priors and show that detailed inputs can be reconstructed. We repurpose the gradient inversion attack, where we optimize towards the generation of poisoned training samples when the corresponding gradients are observed at a deliberately harmful model state.

\section{Design}
\label{sec:method}

\subsection{Threat Model and Assumptions}
\label{sec:threat}

We consider an attacker with white-box access to the model architecture with parameters $\theta$, loss function $\mathcal{L}$, and training data $D=\{(x_i,y_i)\}_{i=1}^n$. This is realistic for a compromised ML training service or an insider with control over the training pipeline, such as data pre-processing. The attacker's goal is to degrade model availability: reducing inference prediction accuracy and expanding the model training time.
We do not assume access to test data during inference. 
Specifically, we distinguish two capabilities:

\begin{enumerate}
  \item \textbf{Order-only attacker (\doa).} The attacker can re-order existing, correctly labelled data during data pre-processing, without changing the training samples, labels, model parameters, or training dynamics.
  \item \textbf{Poisoning attacker (\igdoa/\igska).} The attacker may add synthetic, labelled data to a training set that the learner shuffles. The victim model is initialized independently, and the attacker does not directly alter its parameters.
\end{enumerate}

\subsection{Dynamic Data Ordering}

Let $\theta_0$ be the victim initialization and let $\mathcal{S}\subset D$ be an attacker-visible candidate pool. At ordering step $t$, the attacker has a prefix $C_t=(i_1,\ldots,i_t)$. For each unused candidate $i\in\mathcal{S}$, a temporary model is restored to the same checkpoint and updated sequentially on $C_t\mathbin\Vert i$. Denote the resulting target-layer weights by $W_\ell^{(t,i)}$. The candidates are selected based on the score:
\begin{equation}
  s_t(i)=\mathbf{1}^{\top}W_\ell^{(t,i)}\mathbf{1},
  \qquad i_{t+1}=\arg\min_{i\in\mathcal{S}\setminus C_t}s_t(i).
  \label{eq:doa-score}
\end{equation}
We present the Dynamic Data Ordering (DOA) in Algorithm~\ref{alg:doa}.

\begin{algorithm}[t]
\caption{Dynamic Data Ordering Attack (\doa)}
\label{alg:doa}
\begin{algorithmic}[1]
\REQUIRE Checkpoint $\theta_0$, candidate pool $\mathcal{S}$, layer $\ell$, prefix length $k$
\STATE $C\leftarrow()$
\FOR{$t=0,\ldots,k-1$}
  \STATE $s_{\min}\leftarrow+\infty$, $i^\star\leftarrow\varnothing$
  \FORALL{$i\in\mathcal{S}\setminus C$}
    \STATE Restore $\theta_0$; update on sequence $C\mathbin\Vert i$
    \STATE $s\leftarrow\sum_{a,b}(W_\ell)_{ab}$
    \IF{$s<s_{\min}$}
      \STATE $(s_{\min},i^\star)\leftarrow(s,i)$
    \ENDIF
  \ENDFOR
  \STATE $C\leftarrow C\mathbin\Vert i^\star$
\ENDFOR
\STATE \textbf{return} ordered prefix $C$
\end{algorithmic}
\end{algorithm}

\subsection{Soft-Knockout Weight Construction}

The alternative front-end adapts the soft-knockout construction of~\cite{grosse2020security}. For consecutive affine layers
\begin{equation}
 y=\operatorname{ReLU}\!\left(B\operatorname{ReLU}(Ax+a)+b\right),
 \label{eq:two-layer}
\end{equation}
The attack sorts each weight matrix, concentrates a fraction $r$ of its smallest entries into selected rows of $A$, and places the corresponding entries in selected columns of $B$. Alternating row- and column-wise arrangements ``cross'' the low-weight regions, reducing the probability that positive gradients survive both rectifiers. 
Unlike \doa, SKA directly constructs $\theta^\star$ and therefore presumes control of, or an offline copy of, the model weights during training.

\subsection{Repurposing Gradient Inversion Attacks}

Given an adverse state $\theta^\star$ from \doa or SKA, the back-end instantiates the model at $\theta^\star$. For class $c$, it computes a reference update $g_c^\star$ under the configured unique-class user partition. It then optimizes a synthetic input $\tilde{x}$ to match that update:
\begin{equation}
 \min_{\tilde{x}}\;
 1-\frac{\left\langle \nabla_\theta\mathcal{L}(f_{\theta^\star}(\tilde{x}),c),g_c^\star\right\rangle}
 {\left\|\nabla_\theta\mathcal{L}(f_{\theta^\star}(\tilde{x}),c)\right\|_2
  \left\|g_c^\star\right\|_2}
 +\lambda R(\tilde{x}),
 \label{eq:ig}
\end{equation}
where $R$ denotes the image prior used by the gradient inversion (IG) framework. Equation~\eqref{eq:ig} is the magnitude-invariant objective of gradient inversion~\cite{geiping2020inverting}, evaluated at an intentionally adverse state. The resulting $\tilde{x}$ is assigned label $c$ and inserted into a fresh victim's training set. We repeat the process until we get the desired number of poisoned samples. We call the two attack variants \igdoa and \igska.

\section{Evaluation}

We evaluate whether \doa{} can reduce test accuracy by changing the order of existing training examples, without modifying their contents or labels. We vary the number of ordered examples and measure both the initial accuracy deficit and the deficit remaining after five epochs. We then evaluate whether samples generated by \igdoa{} and \igska{} can degrade the accuracy when the augmented training set is shuffled. For the poisoning attacks, we compare conditions using different poison labels and examine which conditions recover toward the clean baseline within the evaluated training period.

\subsection{Experimental Setup}
\label{sec:setup}

\paragraph{Dataset and model.}
We use MNIST for the evaluation, which contains $60{,}000$ training images and $10{,}000$ test images. The classifier is a fully connected network with ReLU after each hidden layer and a softmax output. This provides a proof-of-concept setting for evaluating attacks that target ReLU inactivity. Unless otherwise specified, we train the classifier using PyTorch's Adadelta optimizer with a learning rate of $1.0$ and a fixed random seed of $1$. The figures show test accuracy after the first five epochs. Epoch 0 denotes the end of the first epoch, and epoch 4 denotes the end of the fifth epoch.

\paragraph{\doa{} configuration.}
We sample a candidate pool of $5{,}000$ training examples and greedily construct ordered prefixes of $20$, $50$, $75$, or $100$ examples using the weight-sum objective in \cref{eq:doa-score}. Restricting the candidate pool avoids evaluating all $60{,}000$ examples at every selection step, while varying the prefix length changes the number of examples explicitly selected for adversarial placement. For evaluation, we place the selected prefix before the remaining MNIST training examples. The dataset still contains the same $60{,}000$ examples with their original labels; only their order changes. We compare the resulting learning curves against clean training to measure the effect of this ordering.

\paragraph{Poison configuration.}
For \igdoa{}, the inversion model loads the checkpoint obtained immediately after the $75$-example ordered prefix. For \igska{}, it loads the model state produced by the crossed-weight construction. Using a modified version of the \texttt{breaching} gradient-reconstruction framework~\cite{geiping2020inverting}, we generate synthetic inputs by matching class-specific reference gradients at the corresponding model state. Each single-class condition adds $200$ synthetic examples assigned the same digit label, increasing the training set from $60{,}000$ to $60{,}200$ examples. Thus, the added samples correspond to approximately $0.33\%$ of the original training-set size. We keep this budget fixed across the ten single-class conditions, denoted C0--C9, so that these comparisons do not change the number of added samples. We also evaluate an ``All'' condition containing poisons assigned labels from all ten classes. This adds 2,000 synthetic examples, with 200 assigned to each digit class, increasing the training set to 62,000 examples. Its total poison budget is therefore ten times that of a single-class condition. In every experiment, the victim is initialized separately, and the augmented training set is shuffled before training.

\paragraph{Evaluation metrics.}
We measure classification accuracy on the unmodified MNIST test set, including all ten classes. For each epoch $e$, we also measure the accuracy deficit relative to the corresponding clean baseline, $\Delta_e = A_{\mathrm{clean},e} - A_{\mathrm{attack},e}$, expressed in percentage points. A positive deficit indicates lower accuracy under attack, while a negative deficit indicates higher accuracy than the clean baseline. Comparing these deficits across epochs distinguishes an initial disruption from a loss that remains at the end of the training budget. Each condition is represented by one seeded run, so we report descriptive comparisons without confidence intervals or statistical significance claims. The epoch-$4$ measurements describe performance after five epochs.

\begin{figure}[th!]
    \centering
    \includegraphics[width=0.9\linewidth]{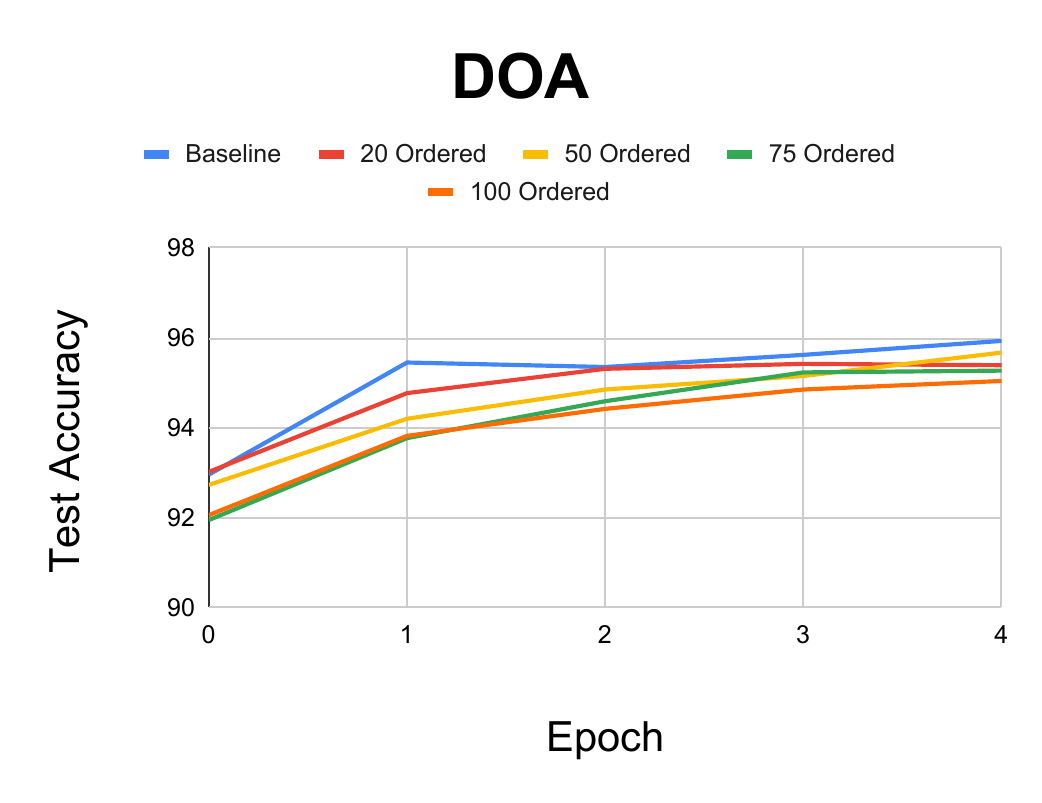}
    \caption{MNIST test accuracy under \doa{} for different ordered-prefix lengths. Epochs $0$--$4$ denote the ends of the first through fifth training epochs. All attacked models improve overall across this period, but remain below the clean baseline at epoch $4$.}
    \label{fig:doa}
\end{figure}

\begin{table}[th!]
    \centering
    \caption{MNIST test accuracy (\%) under \doa{}. The numbered columns
    indicate the number of examples in the ordered prefix.}
    \label{tab:doa}

    \setlength{\tabcolsep}{9pt}
    \begin{tabular}{@{}rrrrrr@{}}
        \toprule
        Epoch & Clean & 20 & 50 & 75 & 100 \\
        \midrule
        0 & 92.96 & 93.02 & 92.73 & 91.95 & 92.06 \\
        1 & 95.45 & 94.77 & 94.20 & 93.77 & 93.82 \\
        2 & 95.35 & 95.31 & 94.85 & 94.59 & 94.42 \\
        3 & 95.62 & 95.42 & 95.15 & 95.23 & 94.85 \\
        4 & 95.93 & 95.39 & 95.67 & 95.27 & 95.04 \\
        \bottomrule
    \end{tabular}
\end{table}

\begin{table}[th!]
    \centering
    \caption{Ordered-prefix size and accuracy deficit relative to clean training. Fractions use the original $60{,}000$ training examples as the denominator. Deficits are in percentage points; positive values indicate lower accuracy under attack. Bold values mark the largest deficit in each epoch column.}
    \label{tab:doa-2}
    \setlength{\tabcolsep}{9pt}
    \begin{tabular}{@{}rrrr@{}}
        \toprule
        \# Ordered & Ordered (\%) & $\Delta_0$ & $\Delta_4$ \\
        \midrule
        0   & 0.000 & 0.00 & 0.00 \\
        20  & 0.033 & $-0.06$ & 0.54 \\
        50  & 0.083 & 0.23 & 0.26 \\
        75  & 0.125 & \textbf{1.01} & 0.66 \\
        100 & 0.167 & 0.90 & \textbf{0.89} \\
        \bottomrule
    \end{tabular}
\end{table}

\subsection{Experimental Results}
\label{sec:results}

\subsubsection{Effect of Data Ordering}
\label{sec:results:doa}

We first measure whether selecting a short adversarial prefix can reduce accuracy when the training examples and labels remain unchanged. \Cref{fig:doa} shows the learning curves, and \cref{tab:doa} reports the accuracy after each epoch. To relate the effect to the attack budget, \cref{tab:doa-2} reports the fraction of training examples selected for the prefix and the accuracy deficits after the first and fifth training epochs.

After the first epoch, the $75$- and $100$-example prefixes reduce test accuracy from $92.96\%$ to $91.95\%$ and $92.06\%$, respectively. 
As training goes further, accuracy drops further. For instance, at epoch 1, test accuracy drops by $1.68\%$ when 75 samples are reordered, and by $1.63\%$ when 100 samples are reordered.
By epoch $4$, all four ordering conditions create an accuracy drop compared to the clean baseline.

Overall, \doa{} reduces accuracy in these runs without changing the training data or labels, but does not prevent the model from continuing to learn. All four attacked models finish with higher accuracy than after their first training epoch. The ordering experiment therefore demonstrates a modest availability effect over the evaluated training period, rather than a complete failure to train.

\subsubsection{Effect of Gradient-Inverted Poisoning}
\label{sec:results:poisoning}

\begin{figure}[th!]
    \centering
    \includegraphics[width=0.88\linewidth]{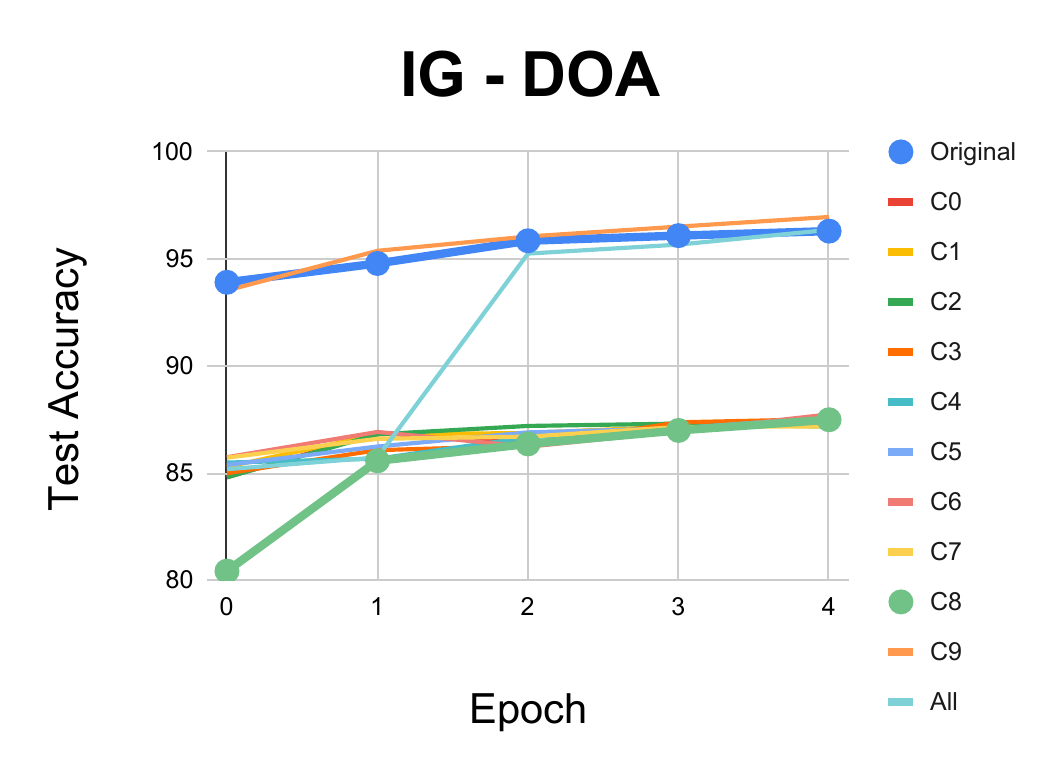}
    \caption{Overall MNIST test accuracy under \igdoa{}. C0--C9 denote
    conditions with $200$ synthetic examples assigned one digit label;
    ``All'' includes poisons from all ten classes. Epochs $0$--$4$
    denote the ends of the first through fifth training epochs.}
    \label{fig:igdoa}
\end{figure}

\begin{figure}[th!]
    \centering
    \includegraphics[width=0.88\linewidth]{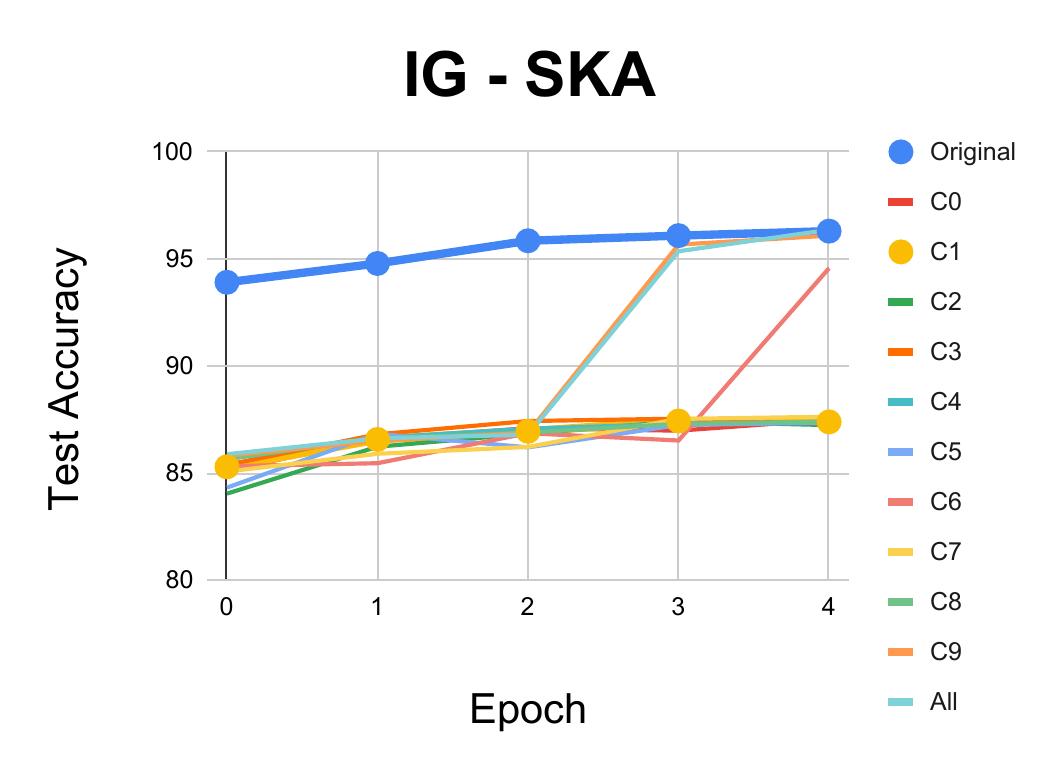}
    \caption{Overall MNIST test accuracy under \igska{}. C0--C9 denote
    conditions with $200$ synthetic examples assigned one digit label;
    ``All'' includes poisons from all ten classes. Epochs $0$--$4$
    denote the ends of the first through fifth training epochs.}
    \label{fig:igska}
\end{figure}

We compare \igdoa{} and \igska{} against the clean baseline shown in Figure~\ref{fig:igdoa} and Figure~\ref{fig:igska} over the five training epochs.

\paragraph{\igdoa{} results.}
As shown in \cref{fig:igdoa}, most single-class conditions begin near $85\%$ accuracy. At epoch $4$, nine of the ten single-class conditions remain at approximately $86$--$88\%$, compared with approximately $96\%$ under clean training. Thus, adding $200$ samples can leave an accuracy deficit of 10\% after five training epochs, even though the learner shuffles the augmented dataset.

C8 is the largest early outlier, beginning near $80\%$ accuracy, approximately $14\%$ below the clean baseline. It improves to approximately $85$--$86\%$ after the next epoch and subsequently approaches the other degraded single-class conditions. 
C9 behaves differently: its accuracy remains close to the clean baseline throughout the evaluated period. The ``All'' condition also reaches approximately clean accuracy, but only after an initial deficit: it begins near $85\%$ and recovers toward the baseline at epoch $2$, the end of the third training epoch. These results show that the generated samples do not have a uniform effect across class conditions. 

\paragraph{\igska{} results.}
As shown in \cref{fig:igska}, eight of the ten single-class conditions remain around $87\%$ accuracy at epoch $4$, approximately $9\%$ below the clean baseline. C9 and ``All'' initially follow the lower-accuracy conditions, but improve sharply at epoch $3$ and finish close to clean accuracy. Unlike C9 under \igdoa{}, C9 under \igska{} therefore causes an initial loss that largely disappears before the end of the training budget. In contrast, C6 accuracy remains around $86$--$87\%$ through epoch $3$, then rises to approximately $94$--$95\%$ at epoch $4$.

Overall, the poisoning experiments demonstrate that gradient-inverted samples can produce accuracy losses that remain substantial after five training epochs, without requiring direct changes to the victim's weights or an attacker-selected presentation order. However, the effect depends on the class condition and the model state used for inversion.

\section{Limitations and Discussion}
\label{sec:limitations}

\paragraph{Evaluation scope and generalization}

Our evaluation demonstrates accuracy degradation on MNIST using a single multilayer perceptron over five training epochs. Each single-class poisoning condition adds $200$ synthetic samples. These experiments establish the observed effects under this configuration; estimating variability across initializations and sensitivity to the poison budget remains future work. Some poisoning conditions recover toward the clean baseline within the evaluated period, while others remain substantially below it. An accuracy deficit after five epochs therefore does not establish that the model cannot recover with further training. Future work could include repeated runs and longer training horizons to characterize the consistency and persistence of the attack’s effects. Evaluation on additional datasets, convolutional networks, and other training configurations could also help determine whether the findings generalize to more diverse settings. Activation choice also warrants investigation. Alternative activations such as leaky ReLU have been used across security and privacy applications, including privacy-policy classification~\cite{qiu2023calpric}, behavioral fingerprint authentication~\cite{wu2020finauth}, privacy-preserving image classification~\cite{li2026ldpkit}, and website-fingerprinting defenses~\cite{holland2024detorrent}. Some of these works explicitly motivate leaky ReLU as a means of sustaining gradient flow or addressing vanishing-gradient problems~\cite{qiu2023calpric,wu2020finauth}, while DeTorrent reports improved performance over standard ReLU~\cite{holland2024detorrent}. These observations motivate evaluating alternative activations, but whether such choices mitigate the degradation observed here remains untested.

\paragraph{Threat model and defenses.}
We assume full knowledge of the model, loss function, and training data, and control over either sample order or the addition of synthetic samples. 
We propose auditing sample order, monitoring activations, and screening for poisons as possible defenses, but do not evaluate them. Their detection accuracy, cost, and ability to reduce the attacks’ effects remain open questions.

\section{Conclusion}
\label{sec:conclusion}

We presented three training-time availability attacks designed to exploit ReLU inactivity through data ordering and poisoning. \doa{} greedily constructs an ordered prefix of existing training examples, while \igdoa{} and \igska{} use gradient inversion to synthesize poisoned samples from model states obtained through data ordering and soft knockout, respectively. In our MNIST evaluation, ordering $100$ of $60{,}000$ training examples reduces test accuracy from $96\%$ to $95\%$ after only five epochs. Adding $200$ poisoned samples leaves most single-class conditions at approximately $86$--$88\%$ accuracy over the same training horizon, compared with approximately $96\%$ under clean training, despite shuffling the augmented dataset and independently initializing the victim. Some conditions recover toward the clean baseline, showing that the effect is not uniform across poison classes. These findings demonstrate that ReLU-targeted attack constructions can impair learning through the training data and their presentation order without directly modifying the victim's parameters. They provide a proof of concept for studying the interaction between activation behavior and training-pipeline integrity.

\section*{Impact Statement}

This work aims to improve the reliability of ML training by studying availability attacks designed to exploit ReLU inactivity through data ordering and poisoning. A potential positive impact is to help researchers and developers identify weaknesses in training pipelines and investigate safeguards, including checks on training-sample order, monitoring of activation and gradient behavior, and detection of poisoned data. These directions could support earlier identification of compromised training runs, reducing the risk of deploying degraded models or spending additional resources on retraining.

The work also presents dual-use risks: publishing the attack methods and implementation could help adversaries manipulate training order or construct poisoned samples to degrade model performance. Our experiments are limited to MNIST in a controlled environment and do not target deployed systems or use private or sensitive data. These restrictions limit direct harm from the experiments but do not eliminate the possibility of downstream misuse. We therefore frame the methods as a basis for reproducible testing and defense development, while making the scope of the evidence explicit: the results demonstrate accuracy degradation in a proof-of-concept setting, not a general vulnerability of deployed ReLU models or the effectiveness of the proposed safeguards.

\begin{credits}
\subsubsection{\ackname}
Funding for this work was provided in part by NSERC Discovery Grant RGPIN-2026-07548. David Lie is supported by a Tier 1 Canada Research Chair (CRC-2019-00242), Wenjun Qiu is supported by the Viola Carless Smith Research Fellowship, and Kexin Li is supported by the Queen Elizabeth II Graduate Scholarship in Science and Technology (QEII-GSST). Kexin Li, Wenjun Qiu, and David Lie are grateful for the support they have received through the Schwartz Reisman Institute for Technology and Society. We also thank Nicolas Papernot, Ilia Shumailov, Mohammad Yaghini, and Jonas Guan for their valuable advice. 

\subsubsection{\discintname}  %
The authors have no competing interests to declare.
\end{credits}

\bibliographystyle{splncs04}
\bibliography{references}

\end{document}